\pdfoutput=1

\documentclass[11pt]{article}

\usepackage[review]{acl}

\usepackage{times}
\usepackage{latexsym}
\usepackage{graphicx}
\usepackage[T1]{fontenc}

\usepackage[utf8]{inputenc}

\usepackage{microtype}

\nolinenumbers
\title{IUTEAM1 at MEDIQA-Chat 2023: Is simple fine tuning effective for multi layer summarization of clinical conversations?}
\author{Dhananjay Srivastava \\
  Indiana University Bloomington \\
  \texttt{dsrivast@iu.edu}}

\begin{document}
{\makeatletter\acl@finalcopytrue
  \maketitle
}
\begin{abstract}
\par Clinical conversation summarization has become an important application of Natural language Processing. In this work, we intend to analyze summarization model ensembling approaches, that can be utilized to improve the overall accuracy of the generated medical report called chart note. The work starts with a single summarization model creating the baseline. Then leads to an ensemble of summarization models trained on a separate section of the chart note. This leads to the final approach of passing the generated results to another summarization model in a multi-layer/stage fashion for better coherency of the generated text. Our results indicate that although an ensemble of models specialized in each section produces better results, the multi-layer/stage approach does not improve accuracy. The code for the above paper is available at https://github.com/dhananjay-srivastava/MEDIQA-Chat-2023-iuteam1.git

\end{abstract}

\section{Introduction}

\par With the increasing adoption of Electronic Health Records (EHRs), physicians and other healthcare professionals are spending an increasing amount of time entering data into EHR systems during patient encounters. It has been estimated that physicians spend approximately 16 minutes per encounter entering data into EHRs \cite{overhage-et-al-ehr-time}.

\par This process can be time consuming and lead to physician burnout.\cite{Babbott_Manwell_Brown_Montague_Williams_Schwartz_Hess_Linzer_2014} In addition, the sheer volume of data generated during a patient encounter can make it difficult for physicians to identify and interpret the most relevant information quickly. To address these challenges, AI summarization models are being developed that can automatically extract and summarize the most important information from clinical conversations \cite{zhang-etal-2021-leveraging-pretrained}.

\par These models can be trained on large datasets of clinical conversations to learn to identify important information such as symptoms, diagnoses, medications, and treatment plans. Once trained, these models can be used to automatically generate summaries of clinical conversations. These summaries can be used to generate first drafts of reports, called chart notes, that must be prepared after each encounter with the patient.

\par There are significant challenges in the implementation of these models.\cite{amin-nejad-etal-2020-exploring} Such as the lack of sufficient training data, ethical and regulatory requirements around sensitive medical data, and the use of specialized medical terminology. The limited availability of clinical data due to privacy concerns makes it difficult to gather a diverse dataset to train the models. Moreover, medical jargon and terminology used by healthcare professionals can vary widely depending on the context, making it challenging to develop models that can accurately identify and summarize critical information. 

\par In this work we explore 3 approaches of combining transformer-based summarization models towards identifying an optimal high-level structure of ensembling multiple summarization models for the task.

\section{Background and Prior Art}

\par In view of the challenges discussed in the previous section, choosing the correct model architecture is crucial. The chartnote is a special document and involves multiple sections each with its own distinct style and content.

\par The purpose of this report is to analyze at a high level, given a transformer-based summarization model

\begin{itemize}
    \item How does a single model perform when it tries to generate the entire chart note from the conversation?
    \item Does a concatenation of results from an ensemble of models trained on each section form better chart notes?
    \item Does passing these generated results through another summarization model generate better chart notes?
\end{itemize}

\par Transformer \cite{vaswani2017attention} based architectures have come to dominate the summarization task. An important challenge in clinical conversation summarization is that the input conversations typically do no fit inside the input token limits of standard models like BERT \cite{devlin-etal-2019-bert} and BART \cite{lewis-etal-2020-bart}. To overcome this challenge models such as Longformer \cite{beltagy2020longformer}, Big Bird \cite{NEURIPS2020_c8512d14} and LSG BART\cite{condevaux2022lsg} have been proposed. We choose the LSG BART model as a sample summarization model to analyze our hypothesis on proper choices for ensembling.

\par LSG BART builds upon BART (Bidirectional and Auto-Regressive Transformer) \cite{lewis-etal-2020-bart} which is a variant of the popular Transformer \cite{vaswani2017attention} architecture that combines the power of auto-regressive and denoising auto-encoder training objectives to generate high-quality summaries. However, the primary limitation in using BART is that it can only accept 1024 tokens. To address this issue, \cite{condevaux2022lsg} have introduced a new technique called LSG attention, which can enhance the performance of BART and other summarization models.

\par LSG attention is a combination of three types of attention mechanisms: local attention, sparse attention, and global attention. In local attention, the input sequence is split into multiple non-overlapping blocks, and attention is calculated within and between these blocks. Sparse attention allows each attention head to process different sparse sets of tokens independently, which can improve the computational efficiency of the model. Global attention, similar to the CLS token, uses a global token to calculate attention across the entire input sequence. Thus this particular model should be suitable for our use case of long document summarization.

\section{Dataset and Challenge Details}

The MEDIQA-Chat 2023 challenge is a part of the 5th Clinical NLP Workshop at ACL 2023 \cite{aci-demo} on improving NLP technology for clinical applications.  The challenge has 3 subtasks. Task A \cite{mts-dialog} is focused on generating specific sections while Task B \cite{ben-abacha-etal-2021-overview} is focused on generating the full note based on the conversation. Task C is focused on generating the conversation back from the note. The Dataset for Task B \cite{ben-abacha-etal-2021-overview} consisted of a Training and Validation Set with 67 and 20 conversations and their summaries respectively. An additional hidden test set of 40 conversations was released to the participants and the final results were calculated using the ROUGE, BLEURT and BERTScore metrics by the competition organizers. This work is focused on Task B.

\section{Methods}

\begin{figure*}
  \centering
  \includegraphics[width=\textwidth]{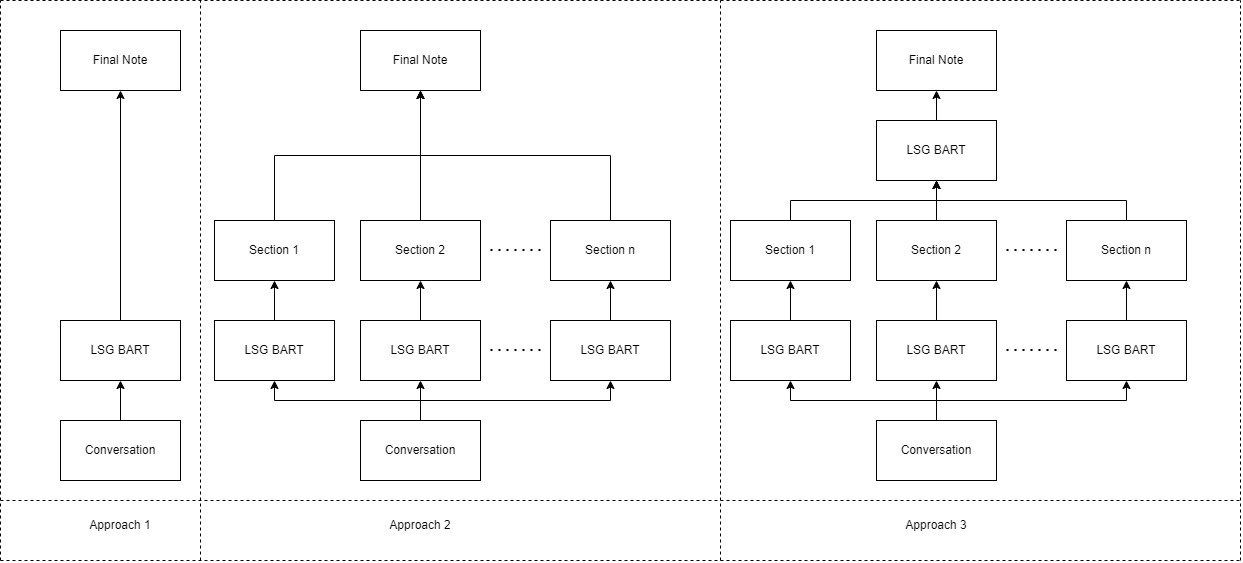}
  \caption{Approach descriptions.}
  \label{fig:approach}
\end{figure*}

\par In order to investigate the hypotheses claimed in the background section, We divide the problem into 3 separate tests using the LSG BART model. We also investigate whether finetuning on Medical Research papers from PUBMED is useful in domain adaptation and whether it leads to an accuracy increase. The approaches are as follows:

\begin{enumerate}
    \item Single LSG BART model with and without finetuning on PubMed Data.
    \item An ensemble of the same LSG BART model but each model is trained on a separate section of the chart note and concatenated to create the final chart note.
    \item A multi-layer model with the first layer being the ensemble of summarizers for each section and another stage/layer of an LSG BART model combining the predictions to create a final chart note.
\end{enumerate}

\subsection{Approach 1}
\par As discussed in the previous sections, the LSG BART model is able to summarize long pieces of text by utilizing Local Sparse and Global Attention mechanisms. A single LSG BART model can accept up to 4096 tokens which are sufficiently large for our input data. Thus we train the dataset on a single model directly which serves as the starting point for our model development and provides a benchmark against which we can compare the performance of other models or modifications to our existing approach.

\subsection{Approach 2}

\par In approach 2 we create an ensemble of summarizers for each separate section of the chart note. The primary motivation for using an ensemble of models is that text internal to a particular section of the chart note is much more coherent than external to it. For e.g., the Content inside the HPI section will contain details about the history of the illness in detail wheras a PE section will contain information regarding any physical exams performed on the patient during the visit.

\par Thus we apply a preprocessing step to the input data to identify and separate different sections within the input text. To achieve this, we have implemented a section extraction script that involves identifying common section headers and grouping them together, for example, "CHIEF COMPLAINT" and "CC" go to the same section "CC". This allows us to extract the relevant information from each section accurately.

\par After the section extraction step is complete, we proceed to train a single LSG BART model for each of the extracted sections. This approach allows us to customize the model training process for each section based on its specific content and characteristics. By doing so, we can optimize the performance of the model for each section and ensure that the resulting summaries are accurate and comprehensive. Once the model training is complete for each section, we concatenate the results to form the final summary of the chart note.

\subsection{Approach 3}
\par For our final approach, we attempt to improve the overall performance of our model by adding another layer/stage of the overall ensemble by passing the generated section texts from Approach 2 into another LSG BART model.

\par The motivation behind this approach is to generate a more coherent and comprehensive summary of the chart note by a better combination of the sections generated in the previous layer/stage. This provides the second LSG BART model with a more complete and diverse set of inputs which should allow the model to observe predictions from different sections and form a more coherent overall summarization.

\section{Results Discussion}

As previously stated, all three approaches in our study utilize the LSG BART model implementation \cite{condevaux2022lsg}. To train the model, we implement a decaying learning rate starting at 5e-5, gradually decreasing the learning rate over time. We train the models for a total of 20 epochs using a single Nvidia A100 GPU and utilize mixed precision training with fp16 set to True for faster training speed with minimal loss to accuracy.

To assess the performance of the models, we evaluate the generated summaries using the ROUGE metric\cite{lin-2004-rouge}, which assesses the degree of overlap between the n-grams in the generated summary and those in the reference summary. The validation set results are as shown in \ref{table:validation_score}

\begin{table}[h!]
\centering
 \begin{tabular}{|c c c c|} 
 \hline
 Approach & Rouge1 & Rouge2 & RougeL \\ 
 \hline
Single BART & 0.497 & 0.241 & 0.264 \\
Section Wise & 0.523 & 0.261 & 0.305 \\
Multilayer & 0.436 & 0.189 & 0.231 \\ 
 \hline
 \end{tabular}
 \caption{Validaton Set Scores for all 3 models}
\label{table:validation_score}
\end{table}

\begin{figure}[h!]
\centering
\includegraphics[scale=0.08]{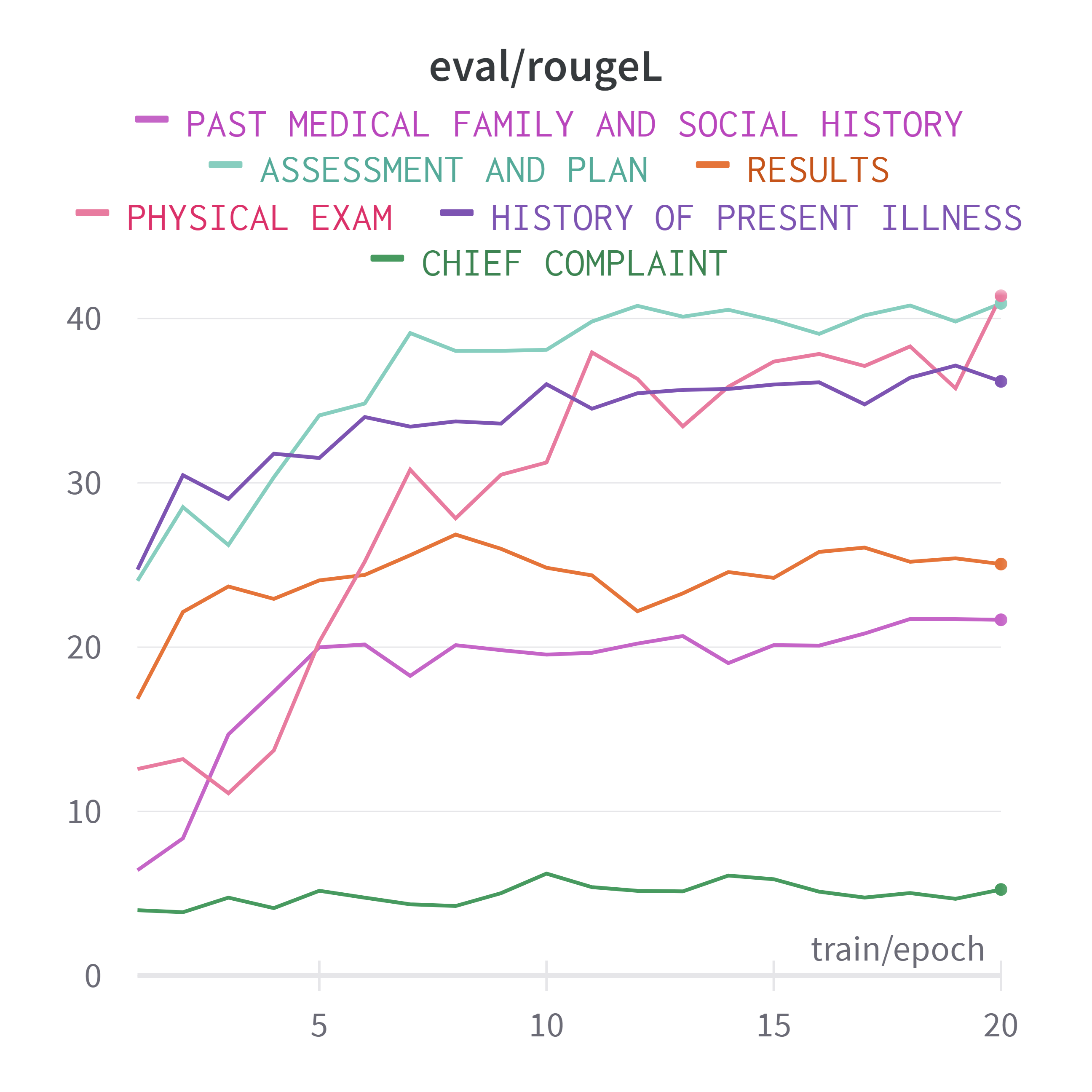}
\caption{ROUGE Scores for Section wise models for LSG BART model finetuned on PubMed Data.}
\label{fig:pubmed}
\end{figure}

 We also utilized finetuned model checkpoints which were trained on medical research papers from PubMed which were further finetuned on our dataset. However as shown in the Rouge score results below, the overall scores are lower for all models finetuned on the PubMed dataset, observe figures \ref{fig: normal} and \ref{fig:pubmed}

\begin{figure}[h!]
\centering
\includegraphics[scale = 0.08]{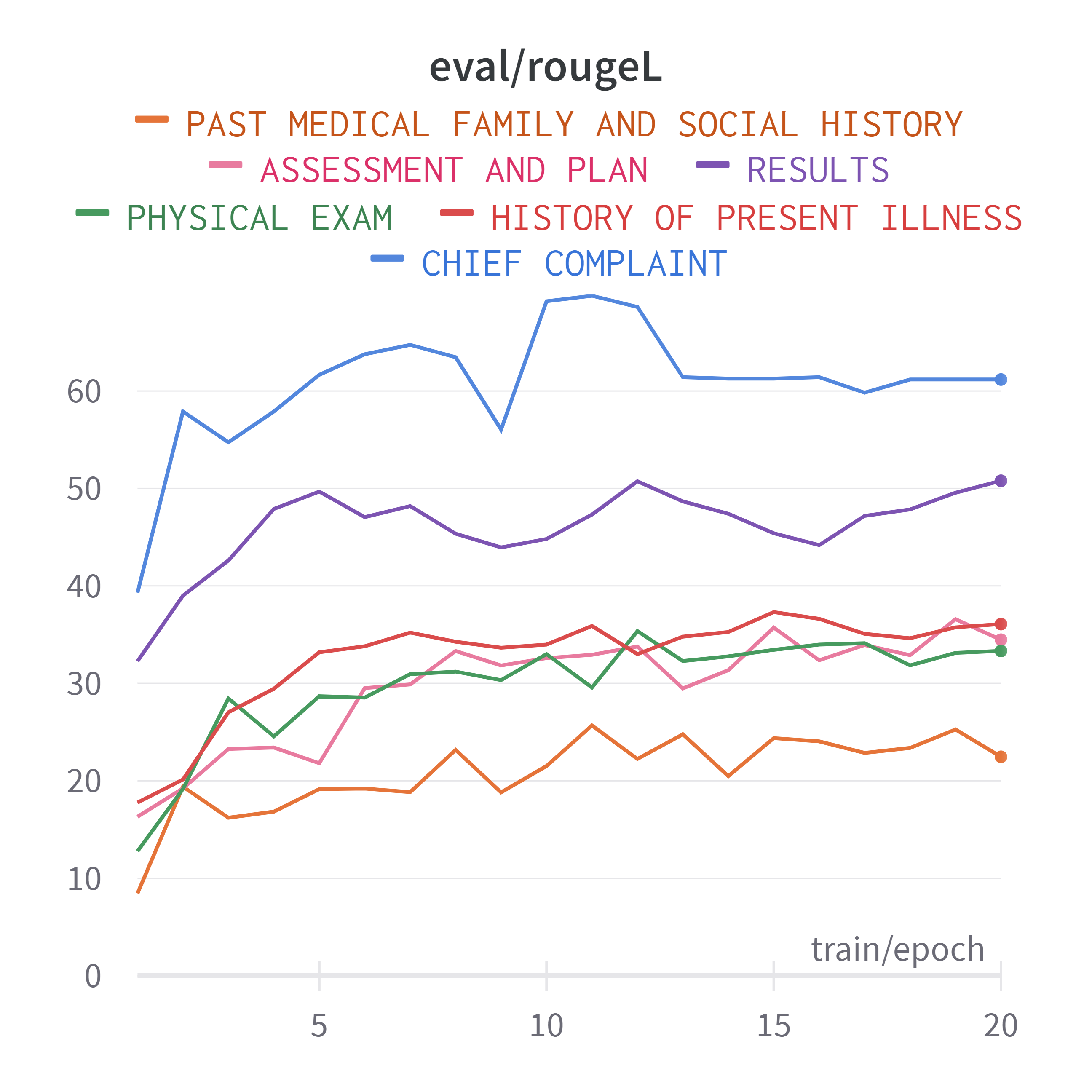}
\caption{ROUGE Scores for Section wise models for LSG BART model.}
\label{fig:normal}
\end{figure}

\par The primary reason behind these lower scores is probably that PubMed data is based on medical literature rather than medical conversations. Moreover, the model was finetuned on this dataset rather than pre-trained thus the model is trained to summarize medical literature but the token embeddings are not necessarily finetuned for our purpose. Thus we did not pursue this model further. 

\par We submitted all the 3 Approaches to MEDIQA-Chat Challenge Task B, The evaluation consisted of 2 parts. In the first part, models from all the different teams were compared on the ROUGE Scores for the full chart note, in the second part the comparison was done sectionwise. The results calculated by the workshop organizers on the hidden test set are.

\begin{table}[h!]
\centering
 \begin{tabular}{|c c c c|} 
 \hline
 Approach & Rouge1 & Rouge2 & RougeL \\ [0.5ex] 
 \hline
Single BART & 0.4917 & 0.2239 & 0.2545 \\
Section Wise & 0.5268 & 0.2622 & 0.306 \\
Multi-Layer & 0.3759 & 0.1786 & 0.2204 \\ [1ex] 
 \hline
 \end{tabular}
 \label{ta}
\end{table}

\par It is observed that Approach 2 seems to work best and has the highest ROUGE scores among the 3 approaches. The results from part 2 help us better understand why this might be happening. 

\begin{table}[h!]
\centering
 \begin{tabular}{|c c c c|}
 \hline
 Approach & Subjective & Exam & Results \\ 
 \hline
Single BART & 0.512	& 0.289 & 0.3525 \\
Section Wise & 0.5456 & 0.5367 & 0.5351 \\
Multi-Layer & 0.5132 & 0.2561 & 0.3848 \\ 
 \hline
 \end{tabular}
 \caption{Section-wise results for the hidden test set.}
\end{table}

\begin{table}[h!]
\centering
 \begin{tabular}{|c c c|}
 \hline
 Approach & Assessment and Plan & Average \\ 
 \hline
Single BART & 0.2842 & 0.3594 \\
Section Wise & 0.5355 & 0.5382 \\
Multi Layer & 0.2424 & 0.3491 \\ 
 \hline
 \end{tabular}
 \caption{Section-wise results for the hidden test set.}
\end{table}

\par We observe that as in approach 2 having an ensemble of models each specializing upon a section of the chart note produces better results than the baseline for all sections. However, attempting to pass the results to another LSG BART model as in approach 3 fails to generate better summaries evidenced by the extensive drop in accuracy for assessment and Plan and Exam sections. Thus model coherency is not improved by simple fine-tuning of a multi-layer/stage summarization ensemble. 

\par In the competition, approach 2 secured 7th and 5th places respectively for full note and section-wise text generation. Approach 1 secured 16th and 17th positions and approach 3 secured 21st and 19th place respectively among the 23 models submitted by the different teams. The competitive ranking of the models and better than baseline performance indicates that ensemble summarization models hold promise and should be investigated further as a viable strategy for clinical conversation summarization.

\section{Conclusion and Future Work}
\par The results indicate that ensembling multiple summarization models depending upon the specific section of the chart note they are producing is a viable strategy for improving summarization quality. Our results also indicate that simply passing the ensemble results to another summarizer does not directly improve accuracy and add that further tests with larger datasets and statistical analyses are required to obtain conclusive answers. In the future, we would like to perform in-depth rigorous analyses on model architectures to support section-wise next generation as well as study many of the other models used in Long Document Summarization to improve overall accuracy.

\section{Ethics Statement}

\par It is important to acknowledge that although the results are promising, language models tend to have hallucinations for generating coherent answers thus these systems should always be used with human supervision. Moreover, this particular system is meant as an experiment to inspire further research into investigating ensembling approaches for summarization and further finetuning as well as model explainability studies are required before they can be used in a clinical setting.

\bibliography{custom}
\bibliographystyle{acl_natbib}

\end{document}